\documentclass[10pt,twocolumn,letterpaper]{article}

\usepackage[pagenumbers]{cvpr}

\usepackage{graphicx}
\usepackage{amsmath}
\usepackage{amssymb}
\usepackage{booktabs}
\usepackage{makecell}

\usepackage{amsmath,amsfonts,bm}

\def\eqref#1{equation~\ref{#1}}

\def\1{\bm{1}}

\DeclareMathAlphabet{\mathsfit}{\encodingdefault}{\sfdefault}{m}{sl}
\SetMathAlphabet{\mathsfit}{bold}{\encodingdefault}{\sfdefault}{bx}{n}

\usepackage{marvosym}
\usepackage[dvipsnames]{xcolor}
\definecolor{cvprblue}{rgb}{0.21,0.49,0.74}
\usepackage[pagebackref,breaklinks,colorlinks,citecolor=cvprblue]{hyperref}

\usepackage[capitalize]{cleveref}
\crefname{section}{Sec.}{Secs.}
\Crefname{section}{Section}{Sections}
\Crefname{table}{Table}{Tables}
\crefname{table}{Tab.}{Tabs.}

\begin{document}

\title{VideoBooth: Diffusion-based Video Generation with Image Prompts}

\author{Yuming Jiang$^{1}$
\quad
Tianxing Wu$^{1}$
\quad
Shuai Yang$^{1}$
\quad
Chenyang Si$^{1}$
\\
Dahua Lin$^{2}$
\quad
Yu Qiao$^{2}$
\quad
Chen Change Loy$^{1}$
\quad
Ziwei Liu\textsuperscript{1\Letter}\\
$^{1}$S-Lab, Nanyang Technological University
\quad
$^{2}$Shanghai AI Laboratory
\\ 
\tt\normalsize\color{Magenta}\url{https://vchitect.github.io/VideoBooth-project/}
}

\twocolumn[{
            \renewcommand\twocolumn[1][]{#1}%
            \vspace{-4em}
            \maketitle
            \begin{center}
                \centering
                \includegraphics[width=0.99\textwidth]{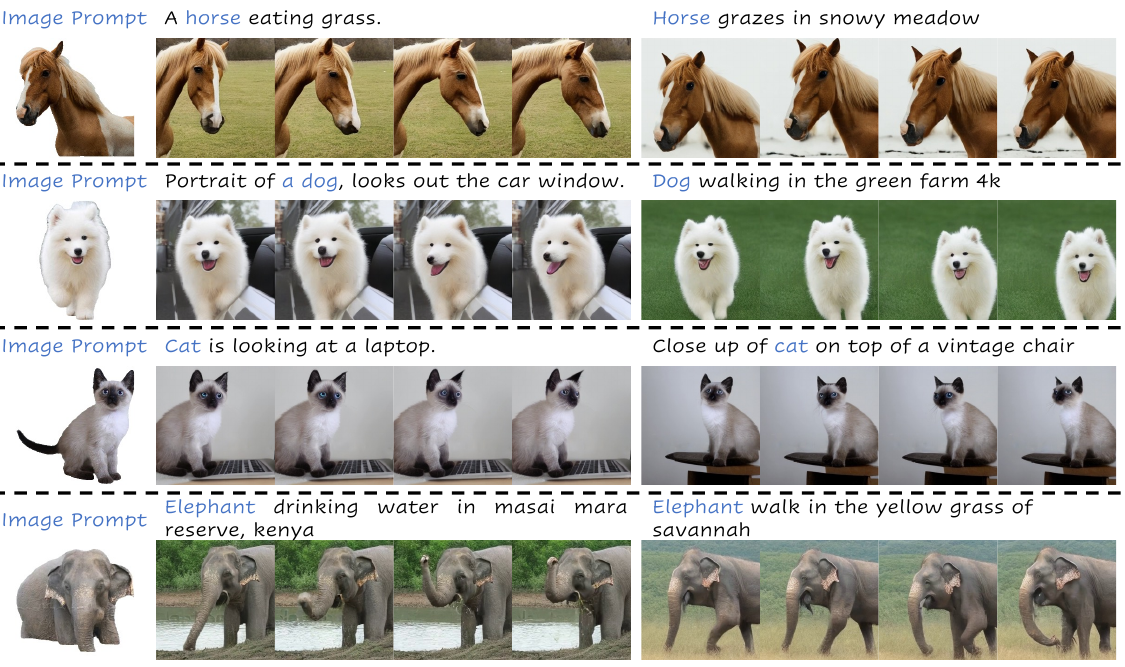}
                \vspace{-10pt}
                \captionof{figure}{\textbf{Videos synthesized by image prompts}. Our VideoBooth generates videos with the subjects specified in the image prompts.}
                \label{teaser}
            \end{center}
        }]


\begin{abstract}
    Text-driven video generation witnesses rapid progress. However, merely using text prompts is not enough to depict the desired subject appearance that accurately aligns with users' intents, especially for customized content creation. In this paper, we study the task of video generation with image prompts, which provide more accurate and direct content control beyond the text prompts. Specifically, we propose a feed-forward framework \textbf{VideoBooth}, with two dedicated designs:
    \textbf{1)} We propose to embed image prompts in a coarse-to-fine manner. 
    Coarse visual embeddings from image encoder provide high-level encodings of image prompts, while fine visual embeddings from the proposed attention injection module provide multi-scale and detailed encoding of image prompts.
    These two complementary embeddings can faithfully capture the desired appearance.
    \textbf{2)} In the attention injection module at fine level, multi-scale image prompts are fed into different cross-frame attention layers as additional keys and values.
    This extra spatial information refines the details in the first frame and then it is propagated to the remaining frames, which maintains temporal consistency.
    Extensive experiments demonstrate that VideoBooth achieves state-of-the-art performance in generating customized high-quality videos with subjects specified in image prompts. Notably, VideoBooth is a generalizable framework where a single model works for a wide range of image prompts with feed-forward pass.
\end{abstract}

\section{Introduction}
\label{sec:intro}

\begin{figure}
   \begin{center}
      \includegraphics[width=0.99\linewidth]{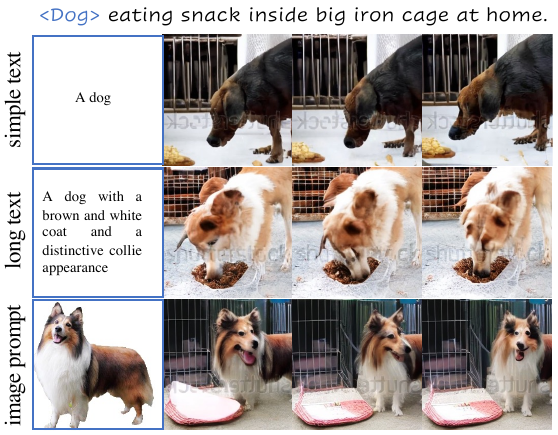}
   \end{center}
  \vspace{-10pt}
   \caption{\textbf{The Use of Image Prompts.} We generate three video clips using different types of prompts: simple text prompt, long text prompt, and image prompt. We use the LLaVa model~\cite{liu2023llava} to generate a text prompt describing the appearance of the image prompt. 
   Using text prompts alone cannot fully capture the visual characteristics of the image prompt.
   }
   \label{necessity_visual}
\end{figure}

Text-to-image models~\cite{ramesh2021zero, ramesh2022hierarchical, saharia2022photorealistic, gafni2022make, ding2022cogview2, ding2021cogview, gu2022vector, rombach2021highresolution, huang2023reversion, si2023freeu, mokady2022null, hertz2022prompt, jiang2022text2human} have attracted substantial attention. With Stable Diffusion~\cite{rombach2021highresolution}, we can now easily generate images using texts. Recently, the focus has been shifted to text-to-video models~\cite{villegas2022phenaki, luo2023videofusion, he2022latent, zhou2022magicvideo, singer2022make, blattmann2023align, esser2023structure, khachatryan2023text2video, wu2022tune, jiang2023text2performer} to generate videos by taking text descriptions as inputs. 
However, in some user cases, texts alone are not expressive enough to define the specific appearance of subjects~\cite{gal2022textual, ruiz2022dreambooth}.
For example, as shown in Fig.~\ref{necessity_visual}, if we want to generate a video clip containing the dog as the third row, we need to use several attributive adjuncts to define the appearance of the dog in text prompts.
Even with these extensive attributive adjuncts, models still cannot generate the desired appearance. 
Defining the appearance of the desired object with texts alone has the following flaws: 1) It is hard to enumerate all desired attributes, and 2) The model cannot capture all attributes accurately with a long text. Compared to using texts, a more straightforward way to define the appearance is to provide reference images, termed image prompts. The image prompts are complementary to text prompts and enrich the details that are hard to be depicted by text prompts.

There are several attempts to introduce image prompts into text-to-image models, which can be roughly divided into two groups. One is to fine-tune parts of parameters using few-shot reference images~\cite{choi2023custom, gal2022textual, ruiz2022dreambooth, kumari2022customdiffusion, han2023svdiff}, which contain the same objects captured under different circumstances. However, the requirement for the number of reference images is demanding as sometimes it is not practical to obtain multiple images of the same object.
The other category~\cite{wei2023elite, jia2023taming, xu2023prompt,xiao2023fastcomposer,li2023blip,chen2023disenbooth}, aiming to address this limitation, proposes to embed image prompts into text-to-image models and the inference is tuning-free.
Both of these two types of attempts achieve plausible results in generating images containing objects specified in image prompts.

In this paper, we study a more challenging task, \ie, text-to-video generation with image prompts. The task has two main challenges: 1) Similar to text-to-image generation, the attributes of image prompts should be accurately captured and then reflected in the generated videos;
2) Different from text-to-image generation, we aim for the dynamic movement of the object rather than a static one.
Directly adapting these methods to video domain results in mismatched appearance or unnatural degraded movements.
To address these challenges, we proposed \textbf{VideoBooth} with elaborately designed coarse-to-fine visual embedding components: 1) Coarse visual embeddings via image encoder: An image encoder is trained to inject the image prompts into text embeddings; 2) Fine visual embeddings via attention injection: The image prompts are mapped to multi-scale latent representations to control the generation process through cross-frame attentions of text-to-video models. 

Specifically, inspired by early attempts~\cite{wei2023elite, jia2023taming} in text-to-image models with image prompts, 
we extract the CLIP image features of the provided image prompts using the pretrained CLIP model~\cite{radford2021learning}. Then the extracted features are mapped into the text embedding space, which are inserted to replace parts of the original text embeddings. The well-trained encoder embeds the coarse appearance information of the given image prompts. 
However, coarse visual embedding is a universal embedding: 1) It only contains high-level semantic information, and 2) It is shared across all blocks with the same scale. As a result, some visual details are missing in the coarse visual embeddings. 

To further refine the generated details as well as maintain temporal consistency, different from the highly compact coarse visual embeddings, multi-scale image prompts are injected into cross-frame attention modules in different layers.
The image prompts provide spatial information as well as details with different granularities.
On the one hand, keeping spatial information of the image prompts can retain more details. On the other hand, different cross-frame attention modules need detailed information at different scales.
Specifically, the latent representations of image prompts are appended as additional keys and values to refine the details in the first generated frame.
To propagate the refined first frame to the following frames to maintain temporal consistency, we then use the updated values of the first frame as values for the remaining frames. 

We set up a dedicated VideoBooth dataset to support the study of the new task. With each video, we provide an image prompt and a text prompt.
Extensive experiments demonstrate the effectiveness of our proposed VideoBooth to generate videos with subjects specified in image prompts. As shown in Fig.~\ref{teaser}, videos generated by VideoBooth better keep the visual attributes of image prompts. Besides, our proposed VideoBooth is tuning-free at inference time and videos can be generated with feed-forward pass only.
The contributions are summarised as follows:
\begin{itemize}
    \item To the best of our knowledge, we are the first to explore the task of video generation using image prompts without finetuning at inference time. We propose a dedicated dataset to support the task.
    Our proposed \textbf{VideoBooth} framework can generate consistent videos containing the subjects specified in image prompts.
    \item We introduce a new coarse-to-fine visual embedding strategy by image encoder and attention injection, which better captures the characteristics of the image prompts.
    \item We propose a novel attention injection method, using the multi-scale image prompts with spatial information to refine the generated details.
\end{itemize}
\section{Related Work}
\label{sec:related}

\noindent \textbf{Text-to-Video Models} take the text descriptions as inputs and generate clips of videos. 
Early explorations~\cite{villegas2022phenaki,hong2022cogvideo} on text-to-video models are based on the idea of VQVAE. 
Make-A-Video~\cite{singer2022make} proposes to add temporal attention to the architecture of DALLE2 model~\cite{ramesh2022hierarchical}. 
Recently, the emergence of diffusion models~\cite{rombach2022high, ho2020denoising} boosts  research on text-to-video models~\cite{luo2023videofusion, zhou2022magicvideo, ge2023pyoco, he2022latent,wang2023lavie,ho2022imagen,an2023latent}.
Video LDM~\cite{blattmann2023align} proposes to train the text-to-video models on Stable Diffusion with temporal attention and 3D convolution introduced to handle the temporal generation.
Gen-1~\cite{esser2023structure} introduces depth maps to handle the temporal consistency of text-to-video models.
Some methods~\cite{guo2023animatediff, zhao2023motiondirector} resort to training separate modules for synthesizing motions.
All of the methods initialize their models with pre-trained text-to-image models.
Another paradigm of using text-to-image models is to directly apply Stable Diffusion~\cite{rombach2021highresolution} to few-shot or zero-shot settings.
Tune-A-Video~\cite{wu2022tune} adapts the self-attention into cross-frame attention and then finetunes the stable diffusion model on a video clip. Models trained in this way have the capability to transfer motions from original videos.
Text2Video-Zero~\cite{khachatryan2023text2video} proposes to generate videos by using correlated noise maps to improve consistency.
Apart from video generation, diffusion models have been applied to video-to-video generations~\cite{qi2023fatezero,shin2023edit,liu2023video,vid2vid-zero,ceylan2023pix2video,zhang2023controlvideo,liew2023magicedit,ouyang2023codef,chai2023stablevideo, yang2023rerender, hu2023videocontrolnet, feng2023ccedit, jeong2023ground, geyer2023tokenflow, yan2023magicprop}.

\noindent \textbf{Customized Content Creation} aims at generating images and videos using reference images~\cite{choi2023custom}.
For customized text-to-image generation, optimization-based methods~\cite{kumari2022customdiffusion, li2023generate, han2023svdiff, hu2022lora} are proposed to optimize the weights of the diffusion model.
For example, Textual Inversion~\cite{gal2022textual} optimizes the word embeddings, while DreamBooth~\cite{ruiz2022dreambooth} proposes to finetune the weights of Stable Diffusion as well.
Optimization-based methods require several reference images with the same subject to avoid the overfitting of the model, which is demanding in real-world applications.
The cost of finetuning hampers the practical usage of these methods.
To address these limitations, encoder-based methods~\cite{xu2023prompt,zhou2023enhancing,ma2023subject,ye2023ip} are proposed to learn a mapping network to embed the reference images.
ELITE~\cite{wei2023elite} proposes to learn a global mapping network and local mapping network to encode the images into word embeddings.
Jia \etal~\cite{jia2023taming} propose to use an additional cross-attention to embed the image features.
With the trained encoder, the personalized generation can be achieved in a feed-forward pass.
Some recent works~\cite{chen2023disenbooth, gong2023talecrafter, gal2023encoder, arar2023domain} combine the encoder-based model and finetuning-based model to improve the performance. BLIP-Diffusion~\cite{li2023blip} proposes to pretrain a multimodal encoder in a large-scale dataset and then finetune the model on the specific subject for inference.
Customized image generation is also applied to place the objects into the user-specified scenes~\cite{chen2023anydoor, bai2023integrating,kulal2023putting, yuan2023customnet,song2023objectstitch}.
Also, some efforts~\cite{yuan2023inserting, chen2023dreamidentity, xiao2023fastcomposer, valevski2023face0, ruiz2023hyperdreambooth, chen2023photoverse, hyung2023magicapture, wu2023singleinsert} have been made to personalized face generation.
He~\etal~\cite{he2023data} propose to improve the performance from the data perspective.
Some works~\cite{liu2023cones,avrahami2023break,jin2023image} focus on composing multiple subjects in one image.
Apart from works on image generation, there are some early attempts at personalized video manipulation.
Make-A-Protagonist~\cite{zhao2023make} edits an existing video in a personalized way using Stable Diffusion 2.1 to embed the image prompts. The motion of the original video is learned from Tune-A-Video~\cite{wu2022tune}.
VideoDreamer~\cite{chen2023videodreamer} proposes to generate personalized videos by generating the first frames using a finetuning-based method and then generating the video clip using the Text2Video-Zero~\cite{khachatryan2023text2video}.
Different from existing works, our proposed VideoBooth does not need to finetune any weights at the inference time.

\section{VideoBooth}
\label{sec:approach}

\begin{figure*}
   \begin{center}
      \includegraphics[width=0.90\linewidth]{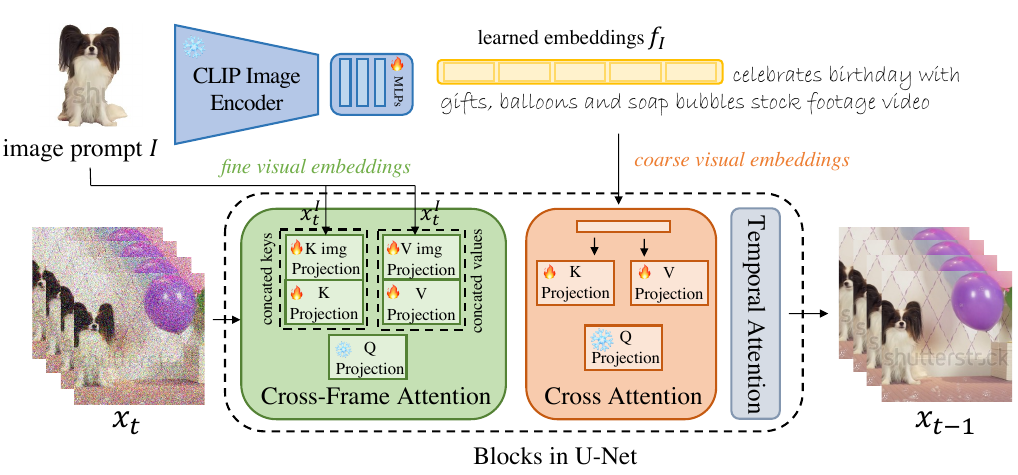}
   \end{center}
  \vspace{-22pt}
   \caption{\textbf{Overview of VideoBooth.} VideoBooth generates videos by taking image prompts $I$ and text prompts $T$ as inputs. The image prompt is fed into the CLIP image encoder, followed by MLP layers. The obtained coarse visual embedding $f_I$ is then inserted into the text embeddings. The composed embeddings serve as the input for cross attention.
   The embedding extracted by the encoder provides a coarse encoding of the visual appearance of the image prompt.
   To further refine the details in the generated videos, at the fine level, we append the latent representation of the image prompt to the cross-frame attention as additional keys and values.
   Different cross-frame attention layers receive latent representations with different scales. The multi-scale features with spatial details refine the synthesized details.
   }
   \label{framework_overview}
\end{figure*}

Our proposed VideoBooth aims at generating videos from an image prompt $I$ and a text prompt $T$. The image prompt specifies the appearance of the subject.
An overview of our proposed VideoBooth is illustrated in Fig.~\ref{framework_overview}.
The image prompt is fed into VideoBooth in two levels.
At the coarse level, it is fed into a pretrained CLIP Image encoder to extract visual features. An encoder, composed of several MLP layers, is trained to map visual features into the space of text embeddings.
The obtained embedding $f_I$ will be inserted into text embedding, which is extracted by feeding text prompt $T$ into CLIP text encoder.
To further refine the synthesized details, we propose to inject image prompt $I$ into the cross-frame attention module in the pretrained video diffusion model.
Specifically, we append latent representation $x^t_I$ of image prompt $I$ into the cross-frame attention. 
In this way, multi-scale visual details with spatial information are involved in the calculation of attention maps so that visual characteristics can be better preserved.
Two ways of feeding image prompt corporate with each other in a coarse-to-fine manner. The encoder provides coarse visual embeddings of the image prompt, while the attention injection provides fine visual embeddings.

\subsection{Preliminary: Pretrained Text-to-Video Model}
Our proposed VideoBooth is developed based on the pretrained text-to-video model~\cite{wang2023lavie, he2022latent}. 
In this section, we will briefly introduce the framework of text-to-video model.

\noindent\textbf{Inflated 2D Conv.}
To handle video data and capture the temporal correlation, 2D conv in the Stable Diffusion model is inflated to 3D conv. In this way, the U-Net can encode 3D features containing the temporal dimension.

\noindent\textbf{Cross-Frame Attention Module.}
Stable Diffusion has a self-attention module, where the features are enhanced by attending to themselves.
To improve temporal consistency, the self-attention is modified into cross-frame attention.
Specifically, in cross-frame attention, the feature of each frame is enhanced by attending and referencing to the first frame and the previous frame. 
The cross-frame attention operates on both the spatial domain and temporal domain, thus the temporal consistency of the synthesized frames is improved.

\noindent\textbf{Temporal Attention Module.}
Apart from cross-frame attention, a temporal attention module is introduced to further improve temporal consistency. 
Temporal attention operates on temporal domain and attends to all frames.

\subsection{Coarse Visual Embeddings via Image Encoder}
Given an image prompt $I$ and text prompt $T$, the generated video is supposed to be consistent with visual elements and textual elements.
Inspired by previous attempts at image-based customization methods~\cite{wei2023elite,jia2023taming}, we propose to encode visual information of image prompts by an image encoder.
The image prompt and text prompt complement each other. 
The image prompt provides visual characteristics of the desired subject in the video, and the text prompt provides other orthogonal information.
The extracted visual embeddings are combined with text embeddings as the final embeddings for the cross-attention module.
Specifically, the CLIP image encoder is employed to extract the visual features $f_V$ of image prompt $I$.
Since the discrepancy exists between the CLIP image and text embeddings, $f_V$ is then fed into MLP layers $F(\cdot)$ to map $f_V$ to the spaces of text embeddings. The final embedding $f_I$ for the image prompt is obtained as follows:
\begin{equation}
    f_V = \text{CLIP}_{I}(I), f_I = F(f_V).
\end{equation}
As for the text prompt $T$, we feed it into the CLIP text encoder to extract the text embedding $f_T$:
\begin{equation}
    f_T = [f^0_T, f^1_T, ..., f^k_T, ...],
\end{equation}
where $f^k_T$ is the $k$-th word embedding in the text prompt.

To make the diffusion model generate videos conditioning on both text prompts and image prompts, we need to integrate these two embeddings, \ie, $f_I$ and $f_T$.
The idea is to replace the word embedding of the target subject with $f_I$.
Mathematically, $f_I$ and $f_t$ are fused to obtain the final text condition $c_t$ as follows:
\begin{equation}
    c_T = [f^0_T, f^1_T, ..., f^{k-1}_T, f_I, f^{k+n}_T, ...],
\end{equation}
where $k$ is the token index of the target subject in the text embedding, and $n$ is the length of the text tokens for the target subject.
For example, an image of a papillon dog is provided as an image prompt. To fuse the information from the image prompt and the text prompt ``Papillon dog celebrates birthday with gifts'', the word embeddings of the ``papillon dog'' will be replaced with $f_I$ before they are fed into the cross-attention module of the diffusion models.

During the training of the coarse stage, we fix the parameters of the CLIP image encoder, and train the MLP layers. 
To make the diffusion model accommodate with the composed text embeddings $c_T$, we also finetune  K and V projections (linear layers to map the input feature to the corresponding keys and values) in the cross-attention module. 

\begin{figure}
   \begin{center}
      \includegraphics[width=0.97\linewidth]{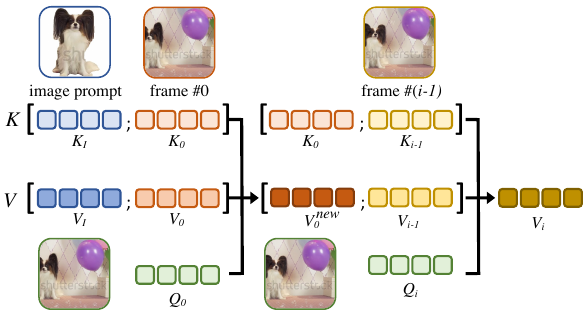}
   \end{center}
  \vspace{-10pt}
   \caption{\textbf{Fine Visual Embedding Refinement.} We propose to inject the latent representation of image prompt (here we use the image for illustration purpose) directly into the cross-frame attention module. We use the keys and values from the image prompt to update the values of the first frame firstly. Then, the updated values of the first frame are used to update the remaining frames. Injecting the image prompt in the cross-frame attention helps to transfer the detailed visual characteristics of the image prompts to the synthesized frames.
   We perform the refinement in different cross-attention layers with different scales. 
   }
  \vspace{-10pt}
   \label{fine_refinement}
\end{figure}

\subsection{Fine Visual Embeddings via Attention Injection}
The well-trained image encoder embeds the coarse visual embeddings for image prompts and thus the synthesized videos contain the subjects specified in image prompts.
However, the image encoder projects the image prompt into a flattened high-level representation, resulting in the loss of its detailed visual cues.
Thus, some detailed visual characteristics in the image prompts may not be well preserved.
To address this problem, a more effective way to preserve these details is to provide the model with the image prompts with spatial resolutions.

To further refine the synthesized details, we propose to inject image prompts into the cross-frame attention of the text-to-video models.
By injecting image prompts into the cross-frame attention, the image prompts are involved in the updates of the synthesized frames so that the model can directly borrow some visual cues from image prompts.

Since text-to-video diffusion models operate in the latent space, %
we first feed the image prompt into the VAE of Stable Diffusion and get its latent representation $x_I$.
Moreover, since the sampling of the videos starts from the noise map, the latent in the intermediate timesteps contains the noises. If we append the clean latent $x_I$ of the image prompt to the cross-frame attention, the domain discrepancy exists. Therefore, we follow the diffusion forward process to add corresponding noises to $x_I$:
\begin{equation}
    x^I_t = \sqrt{\overline{\alpha}_t}x^I_0 + \sqrt{1 - \overline{\alpha}_t}\epsilon,
\end{equation}
where $\overline{\alpha}$ is a hyperparameter determined by the denoising schedule and $\epsilon \sim N(0, I)$. 
 
The cross-frame attention is used to improve the temporal consistency of the generated frames. For each frame, the key and value are the concatenation of the features of the first frame and the previous frame.
Here, we introduce the image prompts as the additional keys and values for the frames.
As shown in Fig.~\ref{fine_refinement}, we propose to update the values of the first frame firstly using the keys and values of the image prompts and the frame itself.
Mathematically, the operation can be expressed as follows:
\begin{equation}
\begin{split}
    V^{new}_0 = softmax(\frac{KQ_0^T}{\sqrt{d}}) \cdot V, \\ 
    K = [K_I, K_0], V = [V_I, V_0],
\end{split}
\end{equation}
where $K_I$ and $V_I$ are the keys and values obtained from the image prompts. The query, key, and value of the first frame are denoted $Q_0$, $K_0$, and $V_0$, respectively. It should be noted that we use a separately trained K and V projection for latent representations $x^I_t$ of image prompts because the image prompts have clean backgrounds, which are different from other frames. The parameters of the newly added K and V projections are initialized by original K and V projections.

Then the updated first frame is used to refine the remaining frames. When updating the remaining frames, the keys used for calculating the attention maps are the original keys, while the values are the updated ones. The update is expressed as follows:
\begin{equation}
\begin{split}
    V^{new}_i = softmax(\frac{KQ_{i}^T}{\sqrt{d}}) \cdot V, \\
    K = [K_0, K_{i-1}], V = [V^{new}_0, V_{i-1}].
\end{split}
\end{equation}

To sum up, in the attention injection, we update the values of the first frame using the image prompts first, and then use the updated first frame to update the other frames. In this way, the visual cues from the image prompts can be consistently propagated to all the frames.

It should be noted that the diffusion model has multiple cross-frame attention layers with different scales. To inject multi-scale visual cues for better detail refinement, in different cross-frame attention layers, we feed latent representations of the image prompts with corresponding resolutions, which are obtained from different stages of the U-Net.

\subsection{Coarse-to-Fine Training Strategy}
The visual details of the image prompts are embedded into the final synthesized results in two stages: coarse visual embeddings using an image encoder and fine visual embedding by attention injection. 
We propose to train these two modules in a coarse-to-fine manner.
In other words, we train the coarse image encoder and tune the parameters in the cross-attention first. After the model has the capability of generating videos containing the subjects specified in image prompts, we then train the attention injection module to embed image prompts into cross-frame attention layers.
As we will show in the ablation study (Sec. \ref{sec:ablation}), if these two modules are trained together, the fine attention injection module leaks the strong visual cues and the coarse encoder learns meaningless representations.
As a result, in sampling phase, the image encoder for the coarse visual embedding cannot provide the coarse information and then the fine attention module cannot refine the details. %
Therefore, it is necessary to train VideoBooth in a coarse-to-fine manner.

\begin{figure*}
   \begin{center}
      \includegraphics[width=0.95\linewidth]{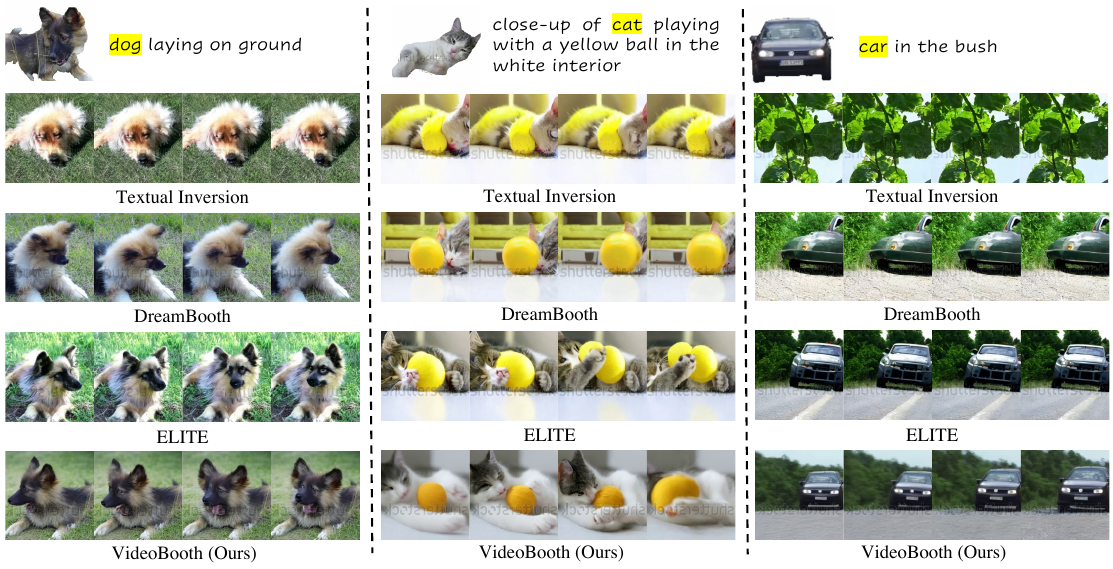}
   \end{center}
  \vspace{-15pt}
   \caption{\textbf{Qualitative Comparison.} VideoBooth effectively preserves the fidelity of image prompts and achieves better visual quality.
   }
  \vspace{-9pt}
   \label{fig_visual_comp}
\end{figure*}

\section{VideoBooth Dataset}

We establish the VideoBooth dataset to support the task of video generation using image prompts. 
We start from the WebVid dataset~\cite{Bain21}, a well-known open-source dataset for text-to-video generation.
In the WebVid dataset, there is a text prompt with each video.
In this paper, we study the task of generating a video clip from one text prompt and one image prompt. 
Hence, in addition to the original text prompt, we need to provide an image prompt for each video.
We propose to segment the subjects from the first frame of the video using the Grounded-SAM (Grounded Segment Anything)~\cite{kirillov2023segany,liu2023grounding}, and the segmented subjects are image prompts. 
The Grounded-SAM receives word prompts as inputs and generates segmentation masks for the target subjects specified in word prompts.
To obtain the word prompt for the input to Grounded-SAM, we use the spaCy library to parse the noun chunks from the original text prompts, which are used as the word prompts.
After the segmentation, we perform data filtering to ensure the data quality. We filter out small objects and large objects (those are almost the same size as the original video) according to the ratio of the object to the whole video. Also, since we focus on generating video clips containing moving objects, we further filter the videos containing moving objects. The keywords we used for filtering are dog, cat, bear, car, panda, tiger, horse, elephant, and lion. 
In the current version, we have processed 2.5M subset of the WebVid dataset.
After data filtering, we have 48,724 video data pairs for training. We will process the full set of the WebVid dataset and include the filtered data in our VideoBooth dataset.

To evaluate the performance, we also set up a test benchmark. 
The test benchmark consists of 650 test pairs. For each pair, an image prompt and a text prompt are provided.
The test pairs are selected from the rest of the WebVid-10M dataset, which does not overlap with the training set.

\section{Experiments}
\label{sec:exp}

\subsection{Comparison Methods}
\noindent\textbf{Textual Inversion}~\cite{gal2022textual} is a method for customized text-to-image generation. The appearance of the target subjects is embedded into the text embeddings. Concretely, a text token $S^*$ is optimized to represent the subject.
We adapt it to the task of text-to-video generation by replacing the image model with the video model.

\noindent\textbf{DreamBooth}~\cite{ruiz2022dreambooth} is also proposed for customized text-to-image generation. It injects the target subject into the text tokens as well as model weights. During the training, both the model weights and the special token $S*$ are optimized.

\noindent\textbf{ELITE}~\cite{wei2023elite} is an encoder-based method for personalized generation. An encoder is trained to embed the images into the text embeddings. Local mapping and global mapping are employed to transform the CLIP embedding of image prompts into the features, which are injected into the cross-attention module. We adapt and retrain the method using the same pretrained video model we use. 

\subsection{Evaluation Metrics}
We use three metrics to evaluate the performance~\cite{wei2023elite}.
To measure the alignment of the generated videos and given text prompts, we use the \textbf{CLIP-Text} metric. The metric is calculated using the cosine similarity of the CLIP text embeddings of text prompts and CLIP image embeddings of the generated frames. For each video, the value is obtained by averaging values of the all frames.
As for the evaluation of the similarity between the given image prompts and the generated videos, we adopt two metrics: \textbf{CLIP-Image} and \textbf{DINO}~\cite{caron2021emerging, oquab2023dinov2}. The CLIP-Image metric is calculated by the cosine similarity between the CLIP image embedding of image prompts and generated frames.
Since the CLIP model is trained to align image embeddings and text embeddings, we follow the practice in previous methods~\cite{ruiz2022dreambooth, chen2023anydoor} and use the DINO similarity as another indicator. DINO is trained to differentiate the differences between objects of the same classes. We use the ViT-S/16 model to extract the features of the image prompts and generated frames. The final score is obtained by averaging over all frames.

\begin{table}
  \centering
    \caption{\textbf{Quantitative Comparisons.} VideoBooth achieves the best image alignment and comparable text alignment performance.}
    \vspace{-5pt}
    \footnotesize{
    \begin{tabular}{l|c|c|c}
    \Xhline{1pt}
    \textbf{Method} & \textbf{CLIP-Text $\uparrow$} & \textbf{CLIP-Image $\uparrow$} & \textbf{DINO $\uparrow$} \\ \Xhline{1pt}
    Textual Inversion~\cite{gal2022textual} & 29.9749 & 69.7995  & 45.3143 \\ 
    DreamBooth~\cite{ruiz2022dreambooth} & \textbf{30.6877} & 71.2078 & 52.9661 \\ 
    ELITE~\cite{wei2023elite} & 30.0881 & 73.7518 & 58.9522 \\ 
    \textbf{VideoBooth (Ours)} & 30.0967 & \textbf{74.7971} & \textbf{65.0979}  \\
    \Xhline{1pt}
  \end{tabular}
  }
  \label{tab:quant_comp}
\end{table}

\subsection{Quantitative Comparisons}
\begin{figure}
   \begin{center}
      \includegraphics[width=1.0\linewidth]{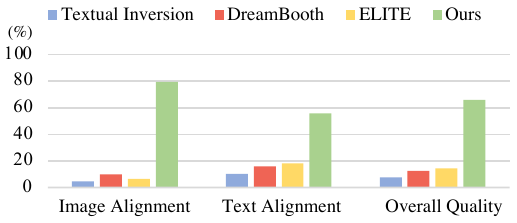}
   \end{center}
  \vspace{-15pt}
   \caption{\textbf{User Study.} Our proposed VideoBooth achieves the highest user preference ratios on all three dimensions.
   }
   \label{fig_user}
\end{figure}

We report quantitative results in Table~\ref{tab:quant_comp}. As shown in Table~\ref{tab:quant_comp}, our proposed VideoBooth achieves state-of-the-art image alignment performance compared to baseline methods. As for the alignment with the text prompts, our proposed VideoBooth has comparable performance with the baseline models. It should be noted that the CLIP-Text score of DreamBooth is significantly higher than the other methods. The reason lies in that the optimized token $S*$ in DreamBooth is inserted into the text embeddings, rather than replacing the original word embeddings like other methods. This would result in the generated videos of DreamBooth being highly related to the text prompts but having no correlation to the image prompts in some cases. 
We also conducted a user study, in which 25 users participated. Each user is presented with twelve groups of videos, and each group contains four videos generated by four methods. For each group, users are asked to make three choices: 1) which one has the best image alignment? 2) which one has the best text alignment? 3) which one has the best overall quality. Figure~\ref{fig_user} summarizes the results. Our results are preferred by most users in all three dimensions.

\begin{figure}
   \begin{center}
      \includegraphics[width=1.0\linewidth]{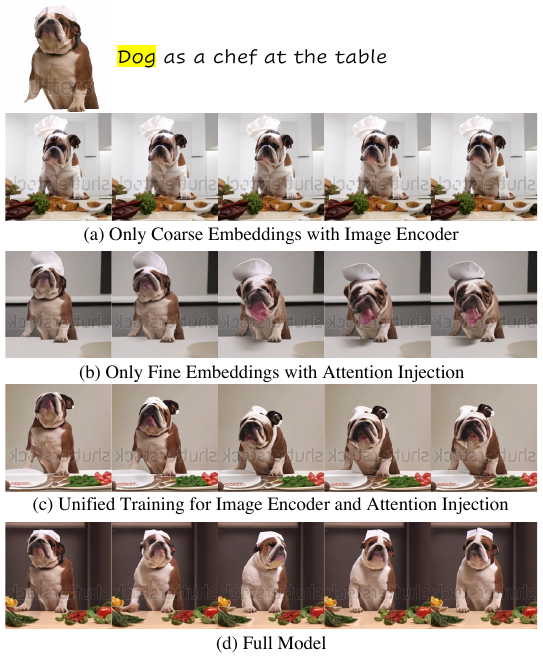}
   \end{center}
  \vspace{-15pt}
   \caption{\textbf{Ablation Study.} (a) With only coarse embeddings from image encoder, generated patterns in the body of the dog are different from the image prompt. (b) With only fine embeddings from attention injection, there lacks the coarse encodings of the dog for the attention injection module to refine and thus the generated dog is distorted at later synthesized frames. (c) The unified training degrades the capability of image encoder and thus the dog is also distorted. (d) The full model better keeps the all visual details of the image prompt.
   }
   \label{fig_ablation}
\end{figure}

\subsection{Qualitative Comparisons}
We show three visual comparisons on the generated video frames of our proposed VideoBooth and baseline methods in Fig.~\ref{fig_visual_comp}. In the first example, the model is supposed to generate videos containing the dog specified in the image laying on the ground. Textual Inversion cannot correctly embed the appearance of the dog and results in generating another totally different dog. DreamBooth and ELITE can embed the coarse appearance of the dog but the generated details vary from the image prompt. Our proposed VideoBooth successfully embeds the details of the image prompts into the synthesized videos.
In the second example, the condition is to generate a cat playing with the yellow ball. All the methods can generate a yellow ball and a cat but only our proposed VideoBooth can accurately generate the cat having the appearance from the image prompt.
As for the last example, Textual Inversion fails to generate a car. DreamBooth generates a distorted car in the bush. 
ELITE model can generate a car in the bush, but the color differs from the image prompt.
By contrast, our model can generate videos having the same car in the bush.

\subsection{Ablation Study}
\label{sec:ablation}
To evaluate the effectiveness of the proposed components, we perform three ablation studies. Due to the computational resources, we train these models on the subset of our training set. 
The quantitative metrics are shown in Table~\ref{tab:ablation}.

\noindent\textbf{Only Coarse Embeddings.}
This ablation model injects the image prompts with only coarse embeddings via Image Encoder. 
In the example shown in Fig.~\ref{fig_ablation}(a), the ablation model only encodes the coarse appearance of the image prompts. The pattern in the legs of the synthesized dog is different from that in the image prompt. By contrast, the results of our full model show that our proposed model can transfer all the details in the image prompts to the synthesized videos.

\noindent\textbf{Only Fine Embeddings.} 
In this ablation model, we only have fine embeddings of the image prompt in cross-frame attention layers.
The main purpose of using fine embedding is to refine the coarse encoding of image prompts from the Image Encoder.
Without coarse embeddings, the model with fine embeddings only cannot refine the details.
As shown in Fig.~\ref{fig_ablation}(b), the first frame contains the exact appearance as the image prompt, but the temporal consistency cannot be guaranteed. The generated dog is distorted in the following frames. The reason is that the model trained in this way overfits the image prompt. In the first frame, the model can copy the information from image prompts. In the following frames, without the coarse embeddings, the generation of the appearance only relies on the propagation of the appearance from the first frame.

\begin{table}
  \centering
    \caption{\textbf{Ablation Study.} The full model has the best scores.}
    \footnotesize{
    \begin{tabular}{l|c|c}
    \Xhline{1pt}
    \textbf{Variants} & \textbf{CLIP-Image $\uparrow$} & \textbf{DINO $\uparrow$} \\ \Xhline{1pt}
    (a) Coarse Embeddings only & 75.4366  & 64.9568 \\ 
    (b) Fine Embeddings only & 75.5553 & 66.0378 \\ 
    (c) Unified Training & 75.8254 & 67.4201 \\ 
    (d) Full Model & \textbf{76.1631} & \textbf{69.7374}  \\
    \Xhline{1pt}
  \end{tabular}
  }
  \label{tab:ablation}
\end{table}

\noindent\textbf{The Necessity of Coarse-to-Fine Training.} 
In VideoBooth, we propose the coarse-to-fine training strategy, \ie, train the coarse embeddings first and then train the attention injection module.
In this ablation model, we train these two modules within one stage.
The unified training makes the model rely heavily on the strong guidance provided in attention injection. In this way, the trained image encoder has limited capability. Thus, the model trained in this way has a similar behavior as the model with only fine embeddings. The model also overfits the image prompts from the attention injection.
As shown in the first example in Fig.~\ref{fig_ablation}(c), the appearance in the first frame is correct, but the generated dog in the following frames is distorted. 
Due to the image encoder having limited capability to provide the correct coarse encoding of image prompts, the attention injection cannot refine the details.
By contrast, the full model can generate consistent frames with all details well preserved.

\section{Discussion}
\label{sec:conclusion}

In this paper, we propose a novel framework VideoBooth to generate videos using image prompts and text prompts. The image prompts specify the appearance of the subjects.
We inject the image prompts into the model in two modules: Coarse Embeddings via Image Encoder and Fine Embeddings via Attention Injection.
The Image Encoder provides the coarse embeddings of image prompts for the refinement of the Attention Injection module. 
These two modules cooperate with each other and they are trained in a coarse-to-fine manner.
Our proposed VideoBooth generates consistent videos containing the desired subjects.

\noindent\textbf{Potential Negative Societal Impacts.}
Our model can be used to synthesize videos. The model may be applied to generate fake videos, which can be potentially avoided by using more advanced fake video detection methods.
%

{
\small
\bibliographystyle{ieeenat_fullname}
\bibliography{egbib}

\begin{thebibliography}{92}
\providecommand{\natexlab}[1]{#1}
\providecommand{\url}[1]{\texttt{#1}}
\expandafter\ifx\csname urlstyle\endcsname\relax
  \providecommand{\doi}[1]{doi: #1}\else
  \providecommand{\doi}{doi: \begingroup \urlstyle{rm}\Url}\fi

\bibitem[An et~al.(2023)An, Zhang, Yang, Gupta, Huang, Luo, and Yin]{an2023latent}
Jie An, Songyang Zhang, Harry Yang, Sonal Gupta, Jia-Bin Huang, Jiebo Luo, and Xi Yin.
\newblock Latent-shift: Latent diffusion with temporal shift for efficient text-to-video generation.
\newblock \emph{arXiv preprint arXiv:2304.08477}, 2023.

\bibitem[Arar et~al.(2023)Arar, Gal, Atzmon, Chechik, Cohen-Or, Shamir, and Bermano]{arar2023domain}
Moab Arar, Rinon Gal, Yuval Atzmon, Gal Chechik, Daniel Cohen-Or, Ariel Shamir, and Amit~H Bermano.
\newblock Domain-agnostic tuning-encoder for fast personalization of text-to-image models.
\newblock \emph{arXiv preprint arXiv:2307.06925}, 2023.

\bibitem[Avrahami et~al.(2023)Avrahami, Aberman, Fried, Cohen-Or, and Lischinski]{avrahami2023break}
Omri Avrahami, Kfir Aberman, Ohad Fried, Daniel Cohen-Or, and Dani Lischinski.
\newblock Break-a-scene: Extracting multiple concepts from a single image.
\newblock \emph{arXiv preprint arXiv:2305.16311}, 2023.

\bibitem[Bai et~al.(2023)Bai, Dong, Feng, Zhang, Ye, Zhou, and Shou]{bai2023integrating}
Jinbin Bai, Zhen Dong, Aosong Feng, Xiao Zhang, Tian Ye, Kaicheng Zhou, and Mike~Zheng Shou.
\newblock Integrating view conditions for image synthesis.
\newblock \emph{arXiv preprint arXiv:2310.16002}, 2023.

\bibitem[Bain et~al.(2021)Bain, Nagrani, Varol, and Zisserman]{Bain21}
Max Bain, Arsha Nagrani, G{\"u}l Varol, and Andrew Zisserman.
\newblock Frozen in time: A joint video and image encoder for end-to-end retrieval.
\newblock In \emph{IEEE International Conference on Computer Vision}, 2021.

\bibitem[Blattmann et~al.(2023)Blattmann, Rombach, Ling, Dockhorn, Kim, Fidler, and Kreis]{blattmann2023align}
Andreas Blattmann, Robin Rombach, Huan Ling, Tim Dockhorn, Seung~Wook Kim, Sanja Fidler, and Karsten Kreis.
\newblock Align your latents: High-resolution video synthesis with latent diffusion models.
\newblock In \emph{Proceedings of the IEEE/CVF Conference on Computer Vision and Pattern Recognition}, pages 22563--22575, 2023.

\bibitem[Caron et~al.(2021)Caron, Touvron, Misra, J{\'e}gou, Mairal, Bojanowski, and Joulin]{caron2021emerging}
Mathilde Caron, Hugo Touvron, Ishan Misra, Herv{\'e} J{\'e}gou, Julien Mairal, Piotr Bojanowski, and Armand Joulin.
\newblock Emerging properties in self-supervised vision transformers.
\newblock In \emph{Proceedings of the IEEE/CVF international conference on computer vision}, pages 9650--9660, 2021.

\bibitem[Ceylan et~al.(2023)Ceylan, Huang, and Mitra]{ceylan2023pix2video}
Duygu Ceylan, Chun-Hao Huang, and Niloy~J. Mitra.
\newblock Pix2video: Video editing using image diffusion.
\newblock \emph{arXiv:2303.12688}, 2023.

\bibitem[Chai et~al.(2023)Chai, Guo, Wang, and Lu]{chai2023stablevideo}
Wenhao Chai, Xun Guo, Gaoang Wang, and Yan Lu.
\newblock Stablevideo: Text-driven consistency-aware diffusion video editing.
\newblock \emph{arXiv preprint arXiv:2308.09592}, 2023.

\bibitem[Chen et~al.(2023{\natexlab{a}})Chen, Wang, Zeng, Zhang, Zhou, Han, and Zhu]{chen2023videodreamer}
Hong Chen, Xin Wang, Guanning Zeng, Yipeng Zhang, Yuwei Zhou, Feilin Han, and Wenwu Zhu.
\newblock Videodreamer: Customized multi-subject text-to-video generation with disen-mix finetuning.
\newblock \emph{arXiv preprint arXiv:2311.00990}, 2023{\natexlab{a}}.

\bibitem[Chen et~al.(2023{\natexlab{b}})Chen, Zhang, Wang, Duan, Zhou, and Zhu]{chen2023disenbooth}
Hong Chen, Yipeng Zhang, Xin Wang, Xuguang Duan, Yuwei Zhou, and Wenwu Zhu.
\newblock Disenbooth: Disentangled parameter-efficient tuning for subject-driven text-to-image generation.
\newblock \emph{arXiv preprint arXiv:2305.03374}, 2023{\natexlab{b}}.

\bibitem[Chen et~al.(2023{\natexlab{c}})Chen, Zhao, Liu, Ding, Song, Wang, Wang, Yang, Liu, Du, et~al.]{chen2023photoverse}
Li Chen, Mengyi Zhao, Yiheng Liu, Mingxu Ding, Yangyang Song, Shizun Wang, Xu Wang, Hao Yang, Jing Liu, Kang Du, et~al.
\newblock Photoverse: Tuning-free image customization with text-to-image diffusion models.
\newblock \emph{arXiv preprint arXiv:2309.05793}, 2023{\natexlab{c}}.

\bibitem[Chen et~al.(2023{\natexlab{d}})Chen, Huang, Liu, Shen, Zhao, and Zhao]{chen2023anydoor}
Xi Chen, Lianghua Huang, Yu Liu, Yujun Shen, Deli Zhao, and Hengshuang Zhao.
\newblock Anydoor: Zero-shot object-level image customization.
\newblock \emph{arXiv preprint arXiv:2307.09481}, 2023{\natexlab{d}}.

\bibitem[Chen et~al.(2023{\natexlab{e}})Chen, Fang, Liu, He, Huang, Zhang, and Mao]{chen2023dreamidentity}
Zhuowei Chen, Shancheng Fang, Wei Liu, Qian He, Mengqi Huang, Yongdong Zhang, and Zhendong Mao.
\newblock Dreamidentity: Improved editability for efficient face-identity preserved image generation.
\newblock \emph{arXiv preprint arXiv:2307.00300}, 2023{\natexlab{e}}.

\bibitem[Choi et~al.(2023)Choi, Choi, Kim, Kim, and Yoon]{choi2023custom}
Jooyoung Choi, Yunjey Choi, Yunji Kim, Junho Kim, and Sungroh Yoon.
\newblock Custom-edit: Text-guided image editing with customized diffusion models.
\newblock \emph{arXiv preprint arXiv:2305.15779}, 2023.

\bibitem[Ding et~al.(2021)Ding, Yang, Hong, Zheng, Zhou, Yin, Lin, Zou, Shao, Yang, et~al.]{ding2021cogview}
Ming Ding, Zhuoyi Yang, Wenyi Hong, Wendi Zheng, Chang Zhou, Da Yin, Junyang Lin, Xu Zou, Zhou Shao, Hongxia Yang, et~al.
\newblock Cogview: Mastering text-to-image generation via transformers.
\newblock \emph{Advances in Neural Information Processing Systems}, 34:\penalty0 19822--19835, 2021.

\bibitem[Ding et~al.(2022)Ding, Zheng, Hong, and Tang]{ding2022cogview2}
Ming Ding, Wendi Zheng, Wenyi Hong, and Jie Tang.
\newblock Cogview2: Faster and better text-to-image generation via hierarchical transformers.
\newblock \emph{arXiv preprint arXiv:2204.14217}, 2022.

\bibitem[Esser et~al.(2023)Esser, Chiu, Atighehchian, Granskog, and Germanidis]{esser2023structure}
Patrick Esser, Johnathan Chiu, Parmida Atighehchian, Jonathan Granskog, and Anastasis Germanidis.
\newblock Structure and content-guided video synthesis with diffusion models.
\newblock \emph{arXiv preprint arXiv:2302.03011}, 2023.

\bibitem[Feng et~al.(2023)Feng, Weng, Wang, Yuan, Bao, Luo, Chen, and Guo]{feng2023ccedit}
Ruoyu Feng, Wenming Weng, Yanhui Wang, Yuhui Yuan, Jianmin Bao, Chong Luo, Zhibo Chen, and Baining Guo.
\newblock Ccedit: Creative and controllable video editing via diffusion models.
\newblock \emph{arXiv preprint arXiv:2309.16496}, 2023.

\bibitem[Gafni et~al.(2022)Gafni, Polyak, Ashual, Sheynin, Parikh, and Taigman]{gafni2022make}
Oran Gafni, Adam Polyak, Oron Ashual, Shelly Sheynin, Devi Parikh, and Yaniv Taigman.
\newblock Make-a-scene: Scene-based text-to-image generation with human priors.
\newblock \emph{arXiv preprint arXiv:2203.13131}, 2022.

\bibitem[Gal et~al.(2022)Gal, Alaluf, Atzmon, Patashnik, Bermano, Chechik, and Cohen-Or]{gal2022textual}
Rinon Gal, Yuval Alaluf, Yuval Atzmon, Or Patashnik, Amit~H. Bermano, Gal Chechik, and Daniel Cohen-Or.
\newblock An image is worth one word: Personalizing text-to-image generation using textual inversion, 2022.

\bibitem[Gal et~al.(2023)Gal, Arar, Atzmon, Bermano, Chechik, and Cohen-Or]{gal2023encoder}
Rinon Gal, Moab Arar, Yuval Atzmon, Amit~H Bermano, Gal Chechik, and Daniel Cohen-Or.
\newblock Encoder-based domain tuning for fast personalization of text-to-image models.
\newblock \emph{ACM Transactions on Graphics (TOG)}, 42\penalty0 (4):\penalty0 1--13, 2023.

\bibitem[Ge et~al.(2023)Ge, Nah, Liu, Poon, Tao, Catanzaro, Jacobs, Huang, Liu, and Balaji]{ge2023pyoco}
Songwei Ge, Seungjun Nah, Guilin Liu, Tyler Poon, Andrew Tao, Bryan Catanzaro, David Jacobs, Jia-Bin Huang, Ming-Yu Liu, and Yogesh Balaji.
\newblock Preserve your own correlation: A noise prior for video diffusion models.
\newblock In \emph{ICCV}, 2023.

\bibitem[Geyer et~al.(2023)Geyer, Bar-Tal, Bagon, and Dekel]{geyer2023tokenflow}
Michal Geyer, Omer Bar-Tal, Shai Bagon, and Tali Dekel.
\newblock Tokenflow: Consistent diffusion features for consistent video editing.
\newblock \emph{arXiv preprint arXiv:2307.10373}, 2023.

\bibitem[Gong et~al.(2023)Gong, Pang, Cun, Xia, Chen, Wang, Zhang, Wang, Shan, and Yang]{gong2023talecrafter}
Yuan Gong, Youxin Pang, Xiaodong Cun, Menghan Xia, Haoxin Chen, Longyue Wang, Yong Zhang, Xintao Wang, Ying Shan, and Yujiu Yang.
\newblock Talecrafter: Interactive story visualization with multiple characters.
\newblock \emph{arXiv preprint arXiv:2305.18247}, 2023.

\bibitem[Gu et~al.(2022)Gu, Chen, Bao, Wen, Zhang, Chen, Yuan, and Guo]{gu2022vector}
Shuyang Gu, Dong Chen, Jianmin Bao, Fang Wen, Bo Zhang, Dongdong Chen, Lu Yuan, and Baining Guo.
\newblock Vector quantized diffusion model for text-to-image synthesis.
\newblock In \emph{Proceedings of the IEEE/CVF Conference on Computer Vision and Pattern Recognition}, pages 10696--10706, 2022.

\bibitem[Guo et~al.(2023)Guo, Yang, Rao, Wang, Qiao, Lin, and Dai]{guo2023animatediff}
Yuwei Guo, Ceyuan Yang, Anyi Rao, Yaohui Wang, Yu Qiao, Dahua Lin, and Bo Dai.
\newblock Animatediff: Animate your personalized text-to-image diffusion models without specific tuning.
\newblock \emph{arXiv preprint arXiv:2307.04725}, 2023.

\bibitem[Han et~al.(2023)Han, Li, Zhang, Milanfar, Metaxas, and Yang]{han2023svdiff}
Ligong Han, Yinxiao Li, Han Zhang, Peyman Milanfar, Dimitris Metaxas, and Feng Yang.
\newblock Svdiff: Compact parameter space for diffusion fine-tuning.
\newblock \emph{arXiv preprint arXiv:2303.11305}, 2023.

\bibitem[He et~al.(2023)He, Cao, Kolkin, Yu, Rhodin, and Kalarot]{he2023data}
Xingzhe He, Zhiwen Cao, Nicholas Kolkin, Lantao Yu, Helge Rhodin, and Ratheesh Kalarot.
\newblock A data perspective on enhanced identity preservation for diffusion personalization.
\newblock \emph{arXiv preprint arXiv:2311.04315}, 2023.

\bibitem[He et~al.(2022)He, Yang, Zhang, Shan, and Chen]{he2022latent}
Yingqing He, Tianyu Yang, Yong Zhang, Ying Shan, and Qifeng Chen.
\newblock Latent video diffusion models for high-fidelity video generation with arbitrary lengths.
\newblock \emph{arXiv preprint arXiv:2211.13221}, 2022.

\bibitem[Hertz et~al.(2022)Hertz, Mokady, Tenenbaum, Aberman, Pritch, and Cohen-Or]{hertz2022prompt}
Amir Hertz, Ron Mokady, Jay Tenenbaum, Kfir Aberman, Yael Pritch, and Daniel Cohen-Or.
\newblock Prompt-to-prompt image editing with cross attention control.
\newblock \emph{arXiv preprint arXiv:2208.01626}, 2022.

\bibitem[Ho et~al.(2020)Ho, Jain, and Abbeel]{ho2020denoising}
Jonathan Ho, Ajay Jain, and Pieter Abbeel.
\newblock Denoising diffusion probabilistic models.
\newblock \emph{Advances in Neural Information Processing Systems}, 33:\penalty0 6840--6851, 2020.

\bibitem[Ho et~al.(2022)Ho, Chan, Saharia, Whang, Gao, Gritsenko, Kingma, Poole, Norouzi, Fleet, et~al.]{ho2022imagen}
Jonathan Ho, William Chan, Chitwan Saharia, Jay Whang, Ruiqi Gao, Alexey Gritsenko, Diederik~P Kingma, Ben Poole, Mohammad Norouzi, David~J Fleet, et~al.
\newblock Imagen video: High definition video generation with diffusion models.
\newblock \emph{arXiv preprint arXiv:2210.02303}, 2022.

\bibitem[Hong et~al.(2022)Hong, Ding, Zheng, Liu, and Tang]{hong2022cogvideo}
Wenyi Hong, Ming Ding, Wendi Zheng, Xinghan Liu, and Jie Tang.
\newblock Cogvideo: Large-scale pretraining for text-to-video generation via transformers.
\newblock \emph{arXiv preprint arXiv:2205.15868}, 2022.

\bibitem[Hu et~al.(2022)Hu, Shen, Wallis, Allen-Zhu, Li, Wang, Wang, and Chen]{hu2022lora}
Edward~J Hu, Yelong Shen, Phillip Wallis, Zeyuan Allen-Zhu, Yuanzhi Li, Shean Wang, Lu Wang, and Weizhu Chen.
\newblock Lo{RA}: Low-rank adaptation of large language models.
\newblock In \emph{International Conference on Learning Representations}, 2022.

\bibitem[Hu and Xu(2023)]{hu2023videocontrolnet}
Zhihao Hu and Dong Xu.
\newblock Videocontrolnet: A motion-guided video-to-video translation framework by using diffusion model with controlnet.
\newblock \emph{arXiv preprint arXiv:2307.14073}, 2023.

\bibitem[Huang et~al.(2023)Huang, Wu, Jiang, Chan, and Liu]{huang2023reversion}
Ziqi Huang, Tianxing Wu, Yuming Jiang, Kelvin~CK Chan, and Ziwei Liu.
\newblock Reversion: Diffusion-based relation inversion from images.
\newblock \emph{arXiv preprint arXiv:2303.13495}, 2023.

\bibitem[Hyung et~al.(2023)Hyung, Shin, and Choo]{hyung2023magicapture}
Junha Hyung, Jaeyo Shin, and Jaegul Choo.
\newblock Magicapture: High-resolution multi-concept portrait customization.
\newblock \emph{arXiv preprint arXiv:2309.06895}, 2023.

\bibitem[Jeong and Ye(2023)]{jeong2023ground}
Hyeonho Jeong and Jong~Chul Ye.
\newblock Ground-a-video: Zero-shot grounded video editing using text-to-image diffusion models.
\newblock \emph{arXiv preprint arXiv:2310.01107}, 2023.

\bibitem[Jia et~al.(2023)Jia, Zhao, Chan, Li, Zhang, Gong, Hou, Wang, and Su]{jia2023taming}
Xuhui Jia, Yang Zhao, Kelvin~CK Chan, Yandong Li, Han Zhang, Boqing Gong, Tingbo Hou, Huisheng Wang, and Yu-Chuan Su.
\newblock Taming encoder for zero fine-tuning image customization with text-to-image diffusion models.
\newblock \emph{arXiv preprint arXiv:2304.02642}, 2023.

\bibitem[Jiang et~al.(2022)Jiang, Yang, Qju, Wu, Loy, and Liu]{jiang2022text2human}
Yuming Jiang, Shuai Yang, Haonan Qju, Wayne Wu, Chen~Change Loy, and Ziwei Liu.
\newblock Text2human: Text-driven controllable human image generation.
\newblock \emph{ACM Transactions on Graphics (TOG)}, 41\penalty0 (4):\penalty0 1--11, 2022.

\bibitem[Jiang et~al.(2023)Jiang, Yang, Koh, Wu, Loy, and Liu]{jiang2023text2performer}
Yuming Jiang, Shuai Yang, Tong~Liang Koh, Wayne Wu, Chen~Change Loy, and Ziwei Liu.
\newblock Text2performer: Text-driven human video generation.
\newblock In \emph{Proceedings of the IEEE/CVF International Conference on Computer Vision}, 2023.

\bibitem[Jin et~al.(2023)Jin, Tanno, Saseendran, Diethe, and Teare]{jin2023image}
Chen Jin, Ryutaro Tanno, Amrutha Saseendran, Tom Diethe, and Philip Teare.
\newblock An image is worth multiple words: Learning object level concepts using multi-concept prompt learning.
\newblock \emph{arXiv preprint arXiv:2310.12274}, 2023.

\bibitem[Khachatryan et~al.(2023)Khachatryan, Movsisyan, Tadevosyan, Henschel, Wang, Navasardyan, and Shi]{khachatryan2023text2video}
Levon Khachatryan, Andranik Movsisyan, Vahram Tadevosyan, Roberto Henschel, Zhangyang Wang, Shant Navasardyan, and Humphrey Shi.
\newblock Text2video-zero: Text-to-image diffusion models are zero-shot video generators.
\newblock \emph{arXiv preprint arXiv:2303.13439}, 2023.

\bibitem[Kirillov et~al.(2023)Kirillov, Mintun, Ravi, Mao, Rolland, Gustafson, Xiao, Whitehead, Berg, Lo, Doll{\'a}r, and Girshick]{kirillov2023segany}
Alexander Kirillov, Eric Mintun, Nikhila Ravi, Hanzi Mao, Chloe Rolland, Laura Gustafson, Tete Xiao, Spencer Whitehead, Alexander~C. Berg, Wan-Yen Lo, Piotr Doll{\'a}r, and Ross Girshick.
\newblock Segment anything.
\newblock \emph{arXiv:2304.02643}, 2023.

\bibitem[Kulal et~al.(2023)Kulal, Brooks, Aiken, Wu, Yang, Lu, Efros, and Singh]{kulal2023putting}
Sumith Kulal, Tim Brooks, Alex Aiken, Jiajun Wu, Jimei Yang, Jingwan Lu, Alexei~A Efros, and Krishna~Kumar Singh.
\newblock Putting people in their place: Affordance-aware human insertion into scenes.
\newblock In \emph{Proceedings of the IEEE/CVF Conference on Computer Vision and Pattern Recognition}, pages 17089--17099, 2023.

\bibitem[Kumari et~al.(2023)Kumari, Zhang, Zhang, Shechtman, and Zhu]{kumari2022customdiffusion}
Nupur Kumari, Bingliang Zhang, Richard Zhang, Eli Shechtman, and Jun-Yan Zhu.
\newblock Multi-concept customization of text-to-image diffusion.
\newblock 2023.

\bibitem[Li et~al.(2023{\natexlab{a}})Li, Li, and Hoi]{li2023blip}
Dongxu Li, Junnan Li, and Steven~CH Hoi.
\newblock Blip-diffusion: Pre-trained subject representation for controllable text-to-image generation and editing.
\newblock \emph{arXiv preprint arXiv:2305.14720}, 2023{\natexlab{a}}.

\bibitem[Li et~al.(2023{\natexlab{b}})Li, Liu, Wen, and Lee]{li2023generate}
Yuheng Li, Haotian Liu, Yangming Wen, and Yong~Jae Lee.
\newblock Generate anything anywhere in any scene.
\newblock \emph{arXiv preprint arXiv:2306.17154}, 2023{\natexlab{b}}.

\bibitem[Liew et~al.(2023)Liew, Yan, Zhang, Xu, and Feng]{liew2023magicedit}
Jun~Hao Liew, Hanshu Yan, Jianfeng Zhang, Zhongcong Xu, and Jiashi Feng.
\newblock Magicedit: High-fidelity and temporally coherent video editing.
\newblock In \emph{arXiv}, 2023.

\bibitem[Liu et~al.(2023{\natexlab{a}})Liu, Li, Wu, and Lee]{liu2023llava}
Haotian Liu, Chunyuan Li, Qingyang Wu, and Yong~Jae Lee.
\newblock Visual instruction tuning.
\newblock In \emph{NeurIPS}, 2023{\natexlab{a}}.

\bibitem[Liu et~al.(2023{\natexlab{b}})Liu, Zeng, Ren, Li, Zhang, Yang, Li, Yang, Su, Zhu, et~al.]{liu2023grounding}
Shilong Liu, Zhaoyang Zeng, Tianhe Ren, Feng Li, Hao Zhang, Jie Yang, Chunyuan Li, Jianwei Yang, Hang Su, Jun Zhu, et~al.
\newblock Grounding dino: Marrying dino with grounded pre-training for open-set object detection.
\newblock \emph{arXiv preprint arXiv:2303.05499}, 2023{\natexlab{b}}.

\bibitem[Liu et~al.(2023{\natexlab{c}})Liu, Zhang, Li, Lin, and Jia]{liu2023video}
Shaoteng Liu, Yuechen Zhang, Wenbo Li, Zhe Lin, and Jiaya Jia.
\newblock Video-p2p: Video editing with cross-attention control.
\newblock \emph{arXiv preprint arXiv:2303.04761}, 2023{\natexlab{c}}.

\bibitem[Liu et~al.(2023{\natexlab{d}})Liu, Zhang, Shen, Zheng, Zhu, Feng, Liu, Zhao, Zhou, and Cao]{liu2023cones}
Zhiheng Liu, Yifei Zhang, Yujun Shen, Kecheng Zheng, Kai Zhu, Ruili Feng, Yu Liu, Deli Zhao, Jingren Zhou, and Yang Cao.
\newblock Cones 2: Customizable image synthesis with multiple subjects.
\newblock \emph{arXiv preprint arXiv:2305.19327}, 2023{\natexlab{d}}.

\bibitem[Luo et~al.(2023)Luo, Chen, Zhang, Huang, Wang, Shen, Zhao, Zhou, and Tan]{luo2023videofusion}
Zhengxiong Luo, Dayou Chen, Yingya Zhang, Yan Huang, Liang Wang, Yujun Shen, Deli Zhao, Jingren Zhou, and Tieniu Tan.
\newblock Videofusion: Decomposed diffusion models for high-quality video generation.
\newblock In \emph{Proceedings of the IEEE/CVF Conference on Computer Vision and Pattern Recognition}, pages 10209--10218, 2023.

\bibitem[Ma et~al.(2023)Ma, Liang, Chen, and Lu]{ma2023subject}
Jian Ma, Junhao Liang, Chen Chen, and Haonan Lu.
\newblock Subject-diffusion: Open domain personalized text-to-image generation without test-time fine-tuning.
\newblock \emph{arXiv preprint arXiv:2307.11410}, 2023.

\bibitem[Mokady et~al.(2022)Mokady, Hertz, Aberman, Pritch, and Cohen-Or]{mokady2022null}
Ron Mokady, Amir Hertz, Kfir Aberman, Yael Pritch, and Daniel Cohen-Or.
\newblock Null-text inversion for editing real images using guided diffusion models.
\newblock \emph{arXiv preprint arXiv:2211.09794}, 2022.

\bibitem[Oquab et~al.(2023)Oquab, Darcet, Moutakanni, Vo, Szafraniec, Khalidov, Fernandez, Haziza, Massa, El-Nouby, et~al.]{oquab2023dinov2}
Maxime Oquab, Timoth{\'e}e Darcet, Th{\'e}o Moutakanni, Huy Vo, Marc Szafraniec, Vasil Khalidov, Pierre Fernandez, Daniel Haziza, Francisco Massa, Alaaeldin El-Nouby, et~al.
\newblock Dinov2: Learning robust visual features without supervision.
\newblock \emph{arXiv preprint arXiv:2304.07193}, 2023.

\bibitem[Ouyang et~al.(2023)Ouyang, Wang, Xiao, Bai, Zhang, Zheng, Zhou, Chen, and Shen]{ouyang2023codef}
Hao Ouyang, Qiuyu Wang, Yuxi Xiao, Qingyan Bai, Juntao Zhang, Kecheng Zheng, Xiaowei Zhou, Qifeng Chen, and Yujun Shen.
\newblock Codef: Content deformation fields for temporally consistent video processing.
\newblock \emph{arXiv preprint arXiv:2308.07926}, 2023.

\bibitem[Qi et~al.(2023)Qi, Cun, Zhang, Lei, Wang, Shan, and Chen]{qi2023fatezero}
Chenyang Qi, Xiaodong Cun, Yong Zhang, Chenyang Lei, Xintao Wang, Ying Shan, and Qifeng Chen.
\newblock Fatezero: Fusing attentions for zero-shot text-based video editing.
\newblock \emph{arXiv preprint arXiv:2303.09535}, 2023.

\bibitem[Radford et~al.(2021)Radford, Kim, Hallacy, Ramesh, Goh, Agarwal, Sastry, Askell, Mishkin, Clark, et~al.]{radford2021learning}
Alec Radford, Jong~Wook Kim, Chris Hallacy, Aditya Ramesh, Gabriel Goh, Sandhini Agarwal, Girish Sastry, Amanda Askell, Pamela Mishkin, Jack Clark, et~al.
\newblock Learning transferable visual models from natural language supervision.
\newblock In \emph{International Conference on Machine Learning}, pages 8748--8763. PMLR, 2021.

\bibitem[Ramesh et~al.(2021)Ramesh, Pavlov, Goh, Gray, Voss, Radford, Chen, and Sutskever]{ramesh2021zero}
Aditya Ramesh, Mikhail Pavlov, Gabriel Goh, Scott Gray, Chelsea Voss, Alec Radford, Mark Chen, and Ilya Sutskever.
\newblock Zero-shot text-to-image generation.
\newblock In \emph{International Conference on Machine Learning}, pages 8821--8831. PMLR, 2021.

\bibitem[Ramesh et~al.(2022)Ramesh, Dhariwal, Nichol, Chu, and Chen]{ramesh2022hierarchical}
Aditya Ramesh, Prafulla Dhariwal, Alex Nichol, Casey Chu, and Mark Chen.
\newblock Hierarchical text-conditional image generation with clip latents.
\newblock \emph{arXiv preprint arXiv:2204.06125}, 2022.

\bibitem[Rombach et~al.(2021)Rombach, Blattmann, Lorenz, Esser, and Ommer]{rombach2021highresolution}
Robin Rombach, Andreas Blattmann, Dominik Lorenz, Patrick Esser, and Björn Ommer.
\newblock High-resolution image synthesis with latent diffusion models, 2021.

\bibitem[Rombach et~al.(2022)Rombach, Blattmann, Lorenz, Esser, and Ommer]{rombach2022high}
Robin Rombach, Andreas Blattmann, Dominik Lorenz, Patrick Esser, and Bj{\"o}rn Ommer.
\newblock High-resolution image synthesis with latent diffusion models.
\newblock In \emph{Proceedings of the IEEE/CVF Conference on Computer Vision and Pattern Recognition}, pages 10684--10695, 2022.

\bibitem[Ruiz et~al.(2022)Ruiz, Li, Jampani, Pritch, Rubinstein, and Aberman]{ruiz2022dreambooth}
Nataniel Ruiz, Yuanzhen Li, Varun Jampani, Yael Pritch, Michael Rubinstein, and Kfir Aberman.
\newblock Dreambooth: Fine tuning text-to-image diffusion models for subject-driven generation.
\newblock 2022.

\bibitem[Ruiz et~al.(2023)Ruiz, Li, Jampani, Wei, Hou, Pritch, Wadhwa, Rubinstein, and Aberman]{ruiz2023hyperdreambooth}
Nataniel Ruiz, Yuanzhen Li, Varun Jampani, Wei Wei, Tingbo Hou, Yael Pritch, Neal Wadhwa, Michael Rubinstein, and Kfir Aberman.
\newblock Hyperdreambooth: Hypernetworks for fast personalization of text-to-image models.
\newblock \emph{arXiv preprint arXiv:2307.06949}, 2023.

\bibitem[Saharia et~al.(2022)Saharia, Chan, Saxena, Li, Whang, Denton, Ghasemipour, Ayan, Mahdavi, Lopes, et~al.]{saharia2022photorealistic}
Chitwan Saharia, William Chan, Saurabh Saxena, Lala Li, Jay Whang, Emily Denton, Seyed Kamyar~Seyed Ghasemipour, Burcu~Karagol Ayan, S~Sara Mahdavi, Rapha~Gontijo Lopes, et~al.
\newblock Photorealistic text-to-image diffusion models with deep language understanding.
\newblock \emph{arXiv preprint arXiv:2205.11487}, 2022.

\bibitem[Shin et~al.(2023)Shin, Kim, Lee, Lee, and Yoon]{shin2023edit}
Chaehun Shin, Heeseung Kim, Che~Hyun Lee, Sang-gil Lee, and Sungroh Yoon.
\newblock Edit-a-video: Single video editing with object-aware consistency.
\newblock \emph{arXiv preprint arXiv:2303.07945}, 2023.

\bibitem[Si et~al.(2023)Si, Huang, Jiang, and Liu]{si2023freeu}
Chenyang Si, Ziqi Huang, Yuming Jiang, and Ziwei Liu.
\newblock Freeu: Free lunch in diffusion u-net.
\newblock \emph{arXiv preprint arXiv:2309.11497}, 2023.

\bibitem[Singer et~al.(2022)Singer, Polyak, Hayes, Yin, An, Zhang, Hu, Yang, Ashual, Gafni, et~al.]{singer2022make}
Uriel Singer, Adam Polyak, Thomas Hayes, Xi Yin, Jie An, Songyang Zhang, Qiyuan Hu, Harry Yang, Oron Ashual, Oran Gafni, et~al.
\newblock Make-a-video: Text-to-video generation without text-video data.
\newblock \emph{arXiv preprint arXiv:2209.14792}, 2022.

\bibitem[Song et~al.(2023)Song, Zhang, Lin, Cohen, Price, Zhang, Kim, and Aliaga]{song2023objectstitch}
Yizhi Song, Zhifei Zhang, Zhe Lin, Scott Cohen, Brian Price, Jianming Zhang, Soo~Ye Kim, and Daniel Aliaga.
\newblock Objectstitch: Object compositing with diffusion model.
\newblock In \emph{Proceedings of the IEEE/CVF Conference on Computer Vision and Pattern Recognition}, pages 18310--18319, 2023.

\bibitem[Valevski et~al.(2023)Valevski, Wasserman, Matias, and Leviathan]{valevski2023face0}
Dani Valevski, Danny Wasserman, Yossi Matias, and Yaniv Leviathan.
\newblock Face0: Instantaneously conditioning a text-to-image model on a face.
\newblock \emph{arXiv preprint arXiv:2306.06638}, 2023.

\bibitem[Villegas et~al.(2022)Villegas, Babaeizadeh, Kindermans, Moraldo, Zhang, Saffar, Castro, Kunze, and Erhan]{villegas2022phenaki}
Ruben Villegas, Mohammad Babaeizadeh, Pieter-Jan Kindermans, Hernan Moraldo, Han Zhang, Mohammad~Taghi Saffar, Santiago Castro, Julius Kunze, and Dumitru Erhan.
\newblock Phenaki: Variable length video generation from open domain textual description.
\newblock \emph{arXiv preprint arXiv:2210.02399}, 2022.

\bibitem[Wang et~al.(2023{\natexlab{a}})Wang, Xie, Liu, Chen, Cao, Wang, and Shen]{vid2vid-zero}
Wen Wang, kangyang Xie, Zide Liu, Hao Chen, Yue Cao, Xinlong Wang, and Chunhua Shen.
\newblock Zero-shot video editing using off-the-shelf image diffusion models.
\newblock \emph{arXiv preprint arXiv:2303.17599}, 2023{\natexlab{a}}.

\bibitem[Wang et~al.(2023{\natexlab{b}})Wang, Chen, Ma, Zhou, Huang, Wang, Yang, He, Yu, Yang, et~al.]{wang2023lavie}
Yaohui Wang, Xinyuan Chen, Xin Ma, Shangchen Zhou, Ziqi Huang, Yi Wang, Ceyuan Yang, Yinan He, Jiashuo Yu, Peiqing Yang, et~al.
\newblock Lavie: High-quality video generation with cascaded latent diffusion models.
\newblock \emph{arXiv preprint arXiv:2309.15103}, 2023{\natexlab{b}}.

\bibitem[Wei et~al.(2023)Wei, Zhang, Ji, Bai, Zhang, and Zuo]{wei2023elite}
Yuxiang Wei, Yabo Zhang, Zhilong Ji, Jinfeng Bai, Lei Zhang, and Wangmeng Zuo.
\newblock Elite: Encoding visual concepts into textual embeddings for customized text-to-image generation.
\newblock \emph{arXiv preprint arXiv:2302.13848}, 2023.

\bibitem[Wu et~al.(2022)Wu, Ge, Wang, Lei, Gu, Hsu, Shan, Qie, and Shou]{wu2022tune}
Jay~Zhangjie Wu, Yixiao Ge, Xintao Wang, Weixian Lei, Yuchao Gu, Wynne Hsu, Ying Shan, Xiaohu Qie, and Mike~Zheng Shou.
\newblock Tune-a-video: One-shot tuning of image diffusion models for text-to-video generation.
\newblock \emph{arXiv preprint arXiv:2212.11565}, 2022.

\bibitem[Wu et~al.(2023)Wu, Yu, Zhu, Wang, and Bai]{wu2023singleinsert}
Zijie Wu, Chaohui Yu, Zhen Zhu, Fan Wang, and Xiang Bai.
\newblock Singleinsert: Inserting new concepts from a single image into text-to-image models for flexible editing.
\newblock \emph{arXiv preprint arXiv:2310.08094}, 2023.

\bibitem[Xiao et~al.(2023)Xiao, Yin, Freeman, Durand, and Han]{xiao2023fastcomposer}
Guangxuan Xiao, Tianwei Yin, William~T Freeman, Fr{\'e}do Durand, and Song Han.
\newblock Fastcomposer: Tuning-free multi-subject image generation with localized attention.
\newblock \emph{arXiv preprint arXiv:2305.10431}, 2023.

\bibitem[Xu et~al.(2023)Xu, Guo, Wang, Huang, Essa, and Shi]{xu2023prompt}
Xingqian Xu, Jiayi Guo, Zhangyang Wang, Gao Huang, Irfan Essa, and Humphrey Shi.
\newblock Prompt-free diffusion: Taking "text" out of text-to-image diffusion models.
\newblock \emph{arXiv preprint arXiv:2305.16223}, 2023.

\bibitem[Xue et~al.(2019)Xue, Chen, Wu, Wei, and Freeman]{xue2019video}
Tianfan Xue, Baian Chen, Jiajun Wu, Donglai Wei, and William~T Freeman.
\newblock Video enhancement with task-oriented flow.
\newblock \emph{International Journal of Computer Vision}, 127:\penalty0 1106--1125, 2019.

\bibitem[Yan et~al.(2023)Yan, Liew, Mai, Lin, and Feng]{yan2023magicprop}
Hanshu Yan, Jun~Hao Liew, Long Mai, Shanchuan Lin, and Jiashi Feng.
\newblock Magicprop: Diffusion-based video editing via motion-aware appearance propagation.
\newblock \emph{arXiv preprint arXiv:2309.00908}, 2023.

\bibitem[Yang et~al.(2023)Yang, Zhou, Liu, and Loy]{yang2023rerender}
Shuai Yang, Yifan Zhou, Ziwei Liu, and Chen~Change Loy.
\newblock Rerender a video: Zero-shot text-guided video-to-video translation.
\newblock \emph{arXiv preprint arXiv:2306.07954}, 2023.

\bibitem[Ye et~al.(2023)Ye, Zhang, Liu, Han, and Yang]{ye2023ip}
Hu Ye, Jun Zhang, Sibo Liu, Xiao Han, and Wei Yang.
\newblock Ip-adapter: Text compatible image prompt adapter for text-to-image diffusion models.
\newblock \emph{arXiv preprint arXiv:2308.06721}, 2023.

\bibitem[Yuan et~al.(2023{\natexlab{a}})Yuan, Cun, Zhang, Li, Qi, Wang, Shan, and Zheng]{yuan2023inserting}
Ge Yuan, Xiaodong Cun, Yong Zhang, Maomao Li, Chenyang Qi, Xintao Wang, Ying Shan, and Huicheng Zheng.
\newblock Inserting anybody in diffusion models via celeb basis.
\newblock \emph{arXiv preprint arXiv:2306.00926}, 2023{\natexlab{a}}.

\bibitem[Yuan et~al.(2023{\natexlab{b}})Yuan, Cao, Wang, Qi, Yuan, and Shan]{yuan2023customnet}
Ziyang Yuan, Mingdeng Cao, Xintao Wang, Zhongang Qi, Chun Yuan, and Ying Shan.
\newblock Customnet: Zero-shot object customization with variable-viewpoints in text-to-image diffusion models.
\newblock \emph{arXiv preprint arXiv:2310.19784}, 2023{\natexlab{b}}.

\bibitem[Zhang et~al.(2023)Zhang, Wei, Jiang, Zhang, Zuo, and Tian]{zhang2023controlvideo}
Yabo Zhang, Yuxiang Wei, Dongsheng Jiang, Xiaopeng Zhang, Wangmeng Zuo, and Qi Tian.
\newblock Controlvideo: Training-free controllable text-to-video generation.
\newblock \emph{arXiv preprint arXiv:2305.13077}, 2023.

\bibitem[Zhao et~al.(2023{\natexlab{a}})Zhao, Gu, Wu, Zhang, Liu, Wu, Keppo, and Shou]{zhao2023motiondirector}
Rui Zhao, Yuchao Gu, Jay~Zhangjie Wu, David~Junhao Zhang, Jiawei Liu, Weijia Wu, Jussi Keppo, and Mike~Zheng Shou.
\newblock Motiondirector: Motion customization of text-to-video diffusion models.
\newblock \emph{arXiv preprint arXiv:2310.08465}, 2023{\natexlab{a}}.

\bibitem[Zhao et~al.(2023{\natexlab{b}})Zhao, Xie, Hong, Li, and Lee]{zhao2023make}
Yuyang Zhao, Enze Xie, Lanqing Hong, Zhenguo Li, and Gim~Hee Lee.
\newblock Make-a-protagonist: Generic video editing with an ensemble of experts.
\newblock \emph{arXiv preprint arXiv:2305.08850}, 2023{\natexlab{b}}.

\bibitem[Zhou et~al.(2022)Zhou, Wang, Yan, Lv, Zhu, and Feng]{zhou2022magicvideo}
Daquan Zhou, Weimin Wang, Hanshu Yan, Weiwei Lv, Yizhe Zhu, and Jiashi Feng.
\newblock Magicvideo: Efficient video generation with latent diffusion models.
\newblock \emph{arXiv preprint arXiv:2211.11018}, 2022.

\bibitem[Zhou et~al.(2023)Zhou, Zhang, Sun, and Xu]{zhou2023enhancing}
Yufan Zhou, Ruiyi Zhang, Tong Sun, and Jinhui Xu.
\newblock Enhancing detail preservation for customized text-to-image generation: A regularization-free approach.
\newblock \emph{arXiv preprint arXiv:2305.13579}, 2023.

\end{thebibliography}
}

\clearpage

\appendix
\section*{Supplementary}
\renewcommand\thesection{\Alph{section}}
\renewcommand\thefigure{A\arabic{figure}}
\renewcommand\thetable{A\arabic{table}}

\section{Comparison methods}
\label{sec_sup:comp}
\noindent\textbf{Texutal Inversion.}
In Textual Inversion~\cite{gal2022textual}, the appearance of target subjects is embedded into the text embeddings. A text token $S^*$ is optimized to represent one specific subject.
When applied to text-to-image models, multiple images containing the same object are required to optimize the text token $S^*$.
In the setting of video generation, we directly use multiple video clips split from the original long video to optimize the text token.
Once optimized, the text token $S^*$ is used to replace the word embeddings of the target subject in the sentence to sample new videos.

\noindent\textbf{DreamBooth.}
In DreamBooth~\cite{ruiz2022dreambooth}, target subjects are injected into text tokens and model weights simultaneously. During the training, both model weights and a special token $S*$ are optimized.
Similar to Textual Inversion, we use the original video and text description to train the model. Multiple video clips sampled from the long video are employed to optimize weights and text token $S*$.
Once trained, the text token $S^*$ is inserted before the word embeddings of the target object to sample new videos.

\noindent\textbf{ELITE.}
Different from Textual Inversion and DreamBooth, ELITE~\cite{wei2023elite} is an encoder-based method for fast customized generation.
An encoder is trained to transform the images into embeddings. Local mapping and global mapping are employed to transform the CLIP embedding of image prompts into the features, which are fed into the cross-attention module.
We adapt the ELITE to video generation. We train the model using the same data and the same base video model.

\section{More Discussions on Ablation Study}
\label{sec_sup:ablation}
In this section, we show one more visual example of the ablation study. In Fig.~7 of the main paper, we show that the model with only coarse embeddings results in imprecise encoding of appearance. Both the model trained with only fine embeddings and the model trained using the unified training strategy overfit to image prompts. In the two examples shown in Fig.~7 of the main paper, the first frames can take the image prompt, but the generated appearance is distorted along the frames.
In Fig.~\ref{ablation_supp}, we discuss another case, which exhibits a different behaviour.

\noindent\textbf{Only Coarse Embeddings.}
This ablation model injects the image prompts with only coarse embeddings via Image Encoder. As shown in Fig.~\ref{ablation_supp}(a), the coarse embeddings provide coarse but not precise guidance. The face of the dog and the shape of the head are not accurately captured. Our full model shown in Fig.~\ref{ablation_supp}(d) can capture the visual details correctly.

\noindent\textbf{Only Fine Embeddings.}
In this ablation model, we only have fine embeddings of image prompts in cross-frame attention layers. Recall that the purpose of fine embeddings is to refine the encoding from coarse levels.
In the example shown in Fig.~\ref{ablation_supp}(b), the first frame does not successfully embed the image prompt. Without coarse embeddings, the generation of the appearance of the dog relies purely on the propagation from the first frame.
The failure in encoding the image prompt into the first frame results in the following frames having random appearances for the dog.

\noindent\textbf{The Necessity of Coarse-to-Fine Training.}
In VideoBooth, we propose the coarse-to-fine training strategy, \ie, train the coarse embeddings first and then train the attention injection module.
This ablation model is trained within one stage.
In the example shown in Fig.~\ref{ablation_supp}(c), the first frame successfully takes the appearance of the image prompt. The model generates a consistent appearance in all frames, but the motion of this generated clip is small and not aligned with the text prompt.
We found that in the case of generating small motions or static frames, the coarse-to-fine training strategy can work well.
However, when it comes to generating large motions as shown in Fig.~7 of the main paper, the appearance will be distorted along frames.

\section{WaterMark Removal Module}
\label{sec_supp:watermark}
\begin{figure*}[h]
    \vspace{-5pt}
   \begin{center}
      \includegraphics[width=0.8\linewidth]{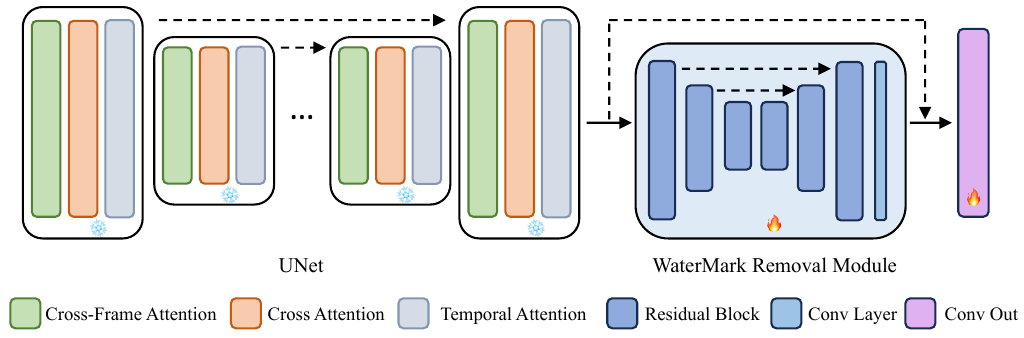}
   \end{center}
  \vspace{-20pt}
   \caption{\textbf{Illustration of Watermark Removal Module.} We add a WaterMark Removal Module before the conv out layer. The output of the watermark removal module is added as a residue to the original features.
   We finetune the newly added module and the conv out layer using the video data ~\cite{xue2019video} without watermarks.}
  \vspace{-10pt}
   \label{watermark}
\end{figure*}

\begin{figure*}[h]
    \vspace{-5pt}
   \begin{center}
      \includegraphics[width=0.8\linewidth]{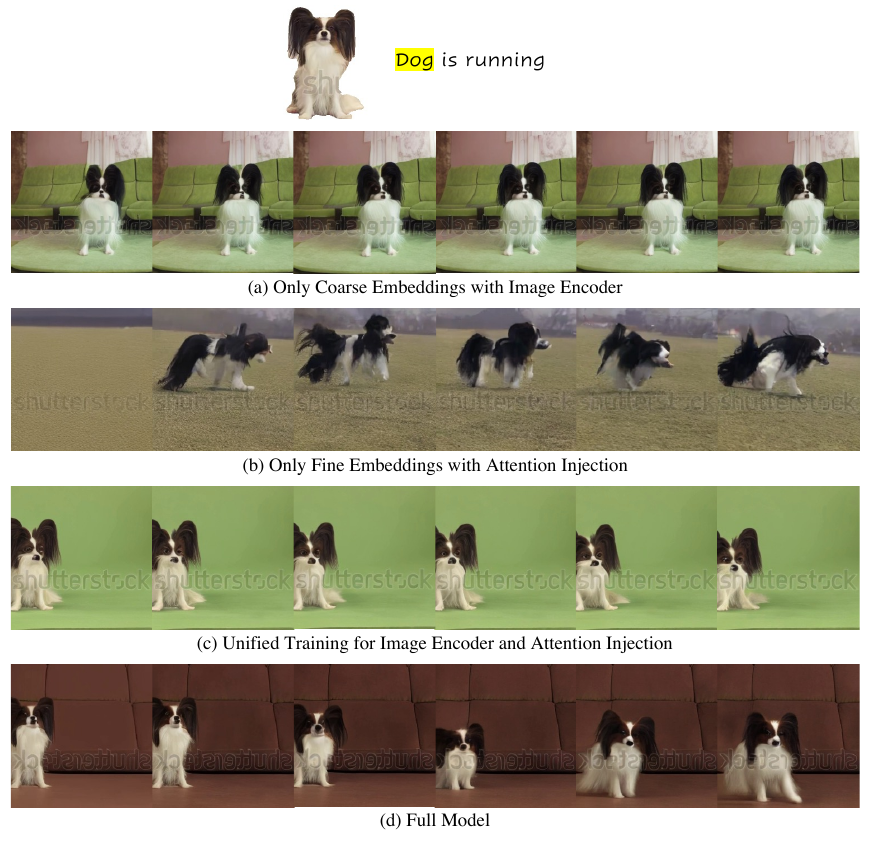}
   \end{center}
  \vspace{-20pt}
   \caption{\textbf{More Visual Analysis on Ablation Study.}}
  \vspace{-15pt}
   \label{ablation_supp}
\end{figure*}

Since the videos in WebVid dataset~\cite{Bain21} have a watermark, the model trained using this dataset generates videos with a watermark in nature.
To generate videos without watermark for better visual quality, we finetune the model with an additional module using the Vimeo dataset~\cite{xue2019video}.
We only use text prompts and original videos to finetune the model.
As shown in Fig.~\ref{watermark}, we add six blocks before the last conv out layer of the base video model.
The added six blocks can be regarded as a small UNet. After the first block, we downsample features by two times. Then after the second block, features are downsampled by two times.
Then features are enhanced by two blocks. Finally, features are upscaled with two consecutive blocks. Each block upsamples features by two times.
Inside each block, there are two ResNet blocks.
Skip connections are adopted between downsampling blocks and upsampling blocks.
After all blocks, we feed the model to one conv layer, which is initialized with zero. The motivation for zero initialization is to avoid the newly added blocks affecting the model.
We add the obtained features to the original features as residues. The added features are fed into the final layer (\ie Conv Out layer) of the base video generation model.
The newly added modules and the last layer are optimized during the finetuning.
After finetuning, the watermark can be removed without influencing the generative capability of VideoBooth.
It should be noted that we use the model without watermark removal module when comparing with baselines.

\newpage

\end{document}